\pdfoutput=1

\documentclass[11pt]{article}

\usepackage[]{EMNLP2023}

\usepackage{times}
\usepackage{latexsym}

\usepackage[T1]{fontenc}

\usepackage[utf8]{inputenc}

\usepackage{microtype}

\usepackage{inconsolata}
\usepackage{times}
\usepackage{color}
\usepackage{latexsym}
\usepackage{amsmath}
\usepackage{booktabs}
\usepackage{multirow}
\usepackage{graphics}
\usepackage{graphicx} 
\usepackage{caption}
\usepackage{amssymb}
\usepackage{pifont}
\usepackage{subfigure}
\usepackage{bm}
\usepackage{bbding}
\usepackage{float}

\newcommand{\model}{GOSE}

\def\mW{{\bm{W}}}
\def\mQ{{\bm{Q}}}
\def\mK{{\bm{K}}}
\def\mV{{\bm{V}}}
\def\mR{{\bm{R}}}

\author{
Xiangnan Chen$^{1}\thanks{~~Work done during an internship at DAMO Research, Alibaba Group.}$ , {\bf Qian Xiao}$^{2}$, Juncheng Li$^{1}\thanks{~~Corresponding author.}$ 
 , Duo Dong$^{1}$, {\bf Jun Lin}$^{2}$,\\ {\bf Xiaozhong Liu}$^{3}$, \textbf{Siliang Tang}$^{1}$\\
 $^{1}$ Zhejiang University 
$^{2}$ Alibaba Group
$^{3}$ Worcester Polytechnic Institute\\
  \texttt{\{xnchen2020,junchengli,22121222,siliang\}@zju.edu.cn}
  \\
  \texttt{\{xiaoqian.xq,linjun.lj\}@alibaba-inc.com, xliu14@wpi.edu}
  } 

\begin{document}

\title{Global Structure Knowledge-Guided Relation Extraction Method for Visually-Rich Document}

\maketitle              

\begin{abstract}
Visual Relation Extraction (VRE) is a powerful means of discovering relationships between entities within visually-rich documents. Existing methods often focus on manipulating entity features to find pairwise relations, yet neglect the more fundamental structural information that links disparate entity pairs together. The absence of global structure information may make the model struggle to learn long-range relations and easily predict conflicted results. To alleviate such limitations, we propose a \textbf{G}l\textbf{O}bal \textbf{S}tructure knowledge-guided relation \textbf{E}xtraction (\textbf{\model}) framework. {\model} initiates by generating preliminary relation predictions on entity pairs extracted from a scanned image of the document. Subsequently, global structural knowledge is captured from the preceding iterative predictions, which are then incorporated into the representations of the entities. This ``generate-capture-incorporate'' cycle is repeated multiple times, allowing entity representations and global structure knowledge to be mutually reinforced. Extensive experiments validate that {\model} not only outperforms existing methods in the standard fine-tuning setting but also reveals superior cross-lingual learning capabilities; indeed, even yields stronger data-efficient performance in the low-resource setting. The code for GOSE will be available at \url{https://github.com/chenxn2020/GOSE}.
\end{abstract}


\section{Introduction}
\label{sec:intro}
Visually-rich document understanding (VrDU) aims to automatically analyze and extract key information from scanned/digital-born documents, such as forms and financial receipts~\cite{funsd,doc-ai}. Since visually-rich documents (VrDs) usually contain diverse structured information, Visual Relation Extraction (VRE), as a critical part of VrDU, has recently attracted extensive attention from both the academic and industrial communities~\cite{funsd,xu2021layoutxlm,spatial,lilt}. The VRE task aims to identify relations between semantic entities in VrDs, severing as the essential basis of mapping VrDs to structured information which is closer to the human comprehension process of the VrDs~\cite{sera}. Recently, inspired by the success of pre-training in visually-rich document understanding~\cite{li2021structurallm,Layoutlm,lilt}, many fine-tuning works predict relation for each entity pair independently, according to local semantic entity representations derived from the pre-trained model.
\begin{figure}
    \centering
    \small
    \includegraphics[width=0.45\textwidth]{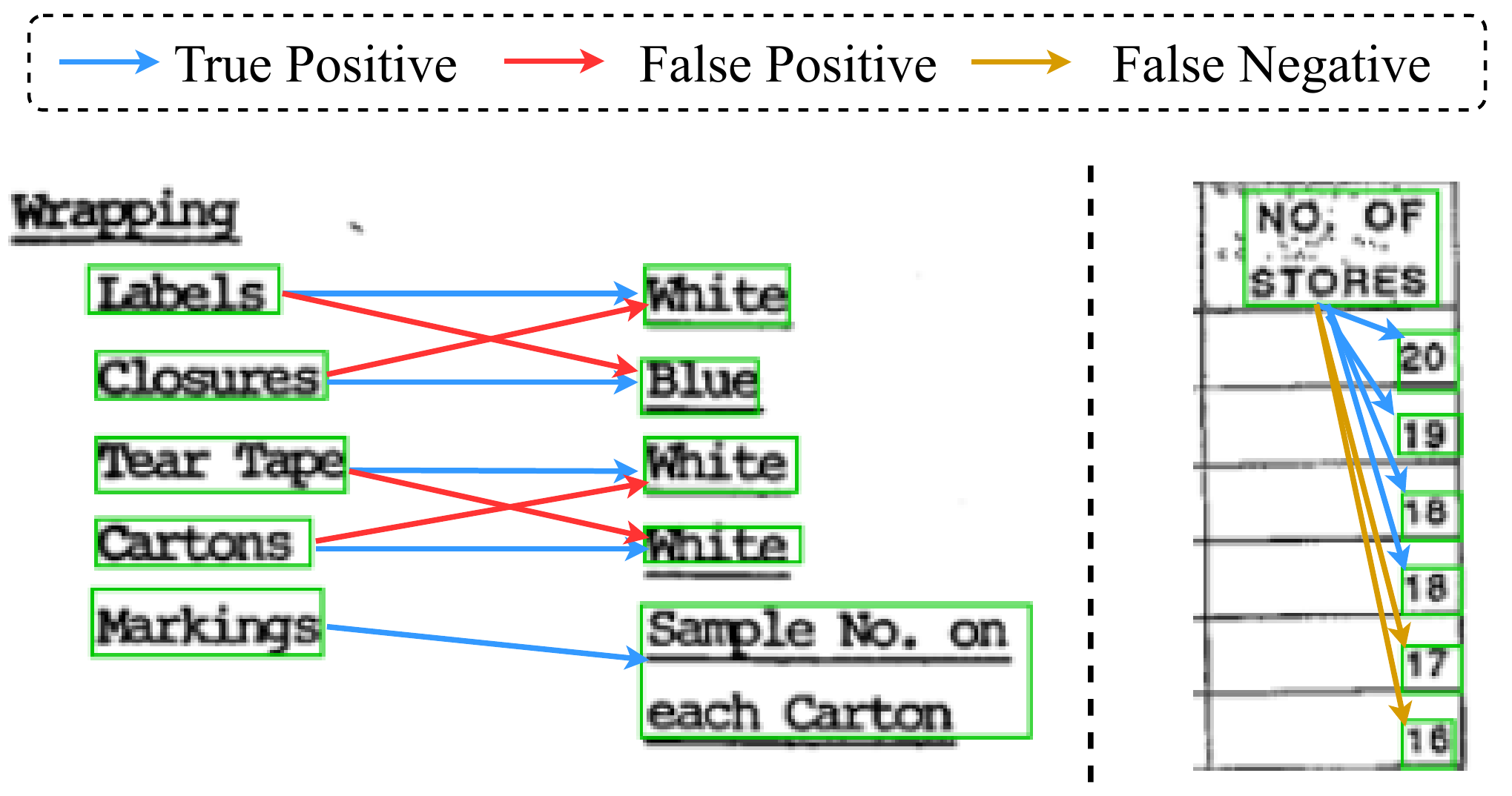} \\
    \centering{\quad\quad\quad\quad(a) Conflicted Links   \quad \quad\quad\quad  (b) Missing Links}
    \caption{Incorrect relation predictions by the LiLT model on FUNSD.}
    \label{fig:example}
    \vspace{-5mm}
\end{figure}



Although existing VRE methods have achieved promising improvement, 
they ignore global structure information, i.e. dependencies between entity pairs. \textbf{Without considering global structure knowledge, the model may easily predict conflicted links and struggle to capture long-range relations.} Taking the state-of-the-art model LiLT~\cite{lilt} as an example, as shown in Figure~\ref{fig:example}(a), although each relational entity pair predicted by the LiLT may make sense semantically, there are conflicted relational entity pairs such as (\textit{Labels}, \textit{Blue}) and (\textit{Closures}, \textit{White}), as well as (\textit{Tear Tape}, \textit{White}) and (\textit{Cartons}, \textit{White}), whose link crosses each other from a global view.
This phenomenon indicates that methods only using local features do not have the sufficient discriminatory ability for conflicted predictions. Furthermore, as shown in Figure \ref{fig:example}(b), even though
LiLT accurately identifies the relational entity pairs (\textit{No.OF STORES}, \textit{19}) and (\textit{No.OF STORES}, \textit{18}), LiLT still hardly learns long-range relations such as (\textit{No.OF STORES}, \textit{16}) and (\textit{No.OF STORES}, \textit{17}). Our intuition is that global structure knowledge can help the model learn long-range relations. The model can predict the relational entity pair (\textit{No.OF STORES}, \textit{17}) by analogy with the global structure consistency between (\textit{No.OF STORES}, \textit{17}) and (\textit{No.OF STORES}, \textit{18}).


In this paper, we present the first study on leveraging global structure information for visual relation extraction. We focus on how to effectively mine and incorporate global structure knowledge into existing fine-tuning methods. 
It has the following two challenges:
(1) \textbf{Huge Mining Space}. Considering {N} entities in a VrD, the computational complexity of capturing dependencies between entity pairs is quadratic to entity pair size ($N^2\times N^2$). So it is difficult to mine useful global structure information in the lack of guidance.
(2) \textbf{Noisy mining process}. Since the process of mining dependencies between entity pairs relies on initial entity representations and lacks direct supervision, global structure information learned by the model is likely to contain noise. Mined noisy global structure knowledge by a model can in turn impair the performance of the model, especially when the model has low prediction accuracy in the early training stage.

To this end, we propose a general global structure knowledge-guided visual relation extraction method named {\model},
which can efficiently and accurately capture dependencies between entity pairs.
{\model} is plug-and-play, which can be flexibly equipped to existing pre-trained VrDU models. Specifically, we first propose a global structure knowledge mining (GSKM) module, which can mine global structure knowledge effectively and efficiently. The GSKM module introduces a novel spatial prefix-guided self-attention mechanism, which takes the spatial layout of entity pairs as the attention prefix to progressively guide mining global structure knowledge in a local-global way. Our intuition is the spatial layout of entity pairs in VrDs may be a valuable clue to uncovering global structure knowledge. As shown in Figure~\ref{fig:example}(a), we can recognize crossover between entity pairs in 2D space by computing the spatial layout of linking lines. Furthermore, in order to increase the robustness of {\model} and handle the noisy mining process, we introduce an iterative learning strategy to combine the process of entity representations learning and global structure mining. The integration of global structure knowledge can help refine entity embeddings, while better entity embeddings can help mine more accurate global structure knowledge. 
In summary, the contributions of our work are as follows: 
\begin{itemize}
     \item We propose a global structure knowledge-guided visual relation extraction method,
    named {\model}. It can use the spatial layout as a clue to mine global structure knowledge effectively and leverage the iterative learning strategy to denoise global structure knowledge.
    \item {\model} can be easily applied to existing pre-trained VrDU models. Experimental results on the standard fine-tuning task over 8 datasets show that our method improves the average F1 performance of the previous SOTA models by a large margin: LiLT(+14.20\%) and LayoutXLM(+12.88\%).
    \item We further perform comprehensive experiments covering diverse settings of VRE tasks, such as cross-lingual learning, and low-resource setting. Experimental results illustrate advantages of our model, such as cross-lingual transfer and data-efficient learning.
\end{itemize}

\section{Preliminaries}
\begin{figure*}[ht]
    \centering
    \includegraphics[width=0.98\textwidth]{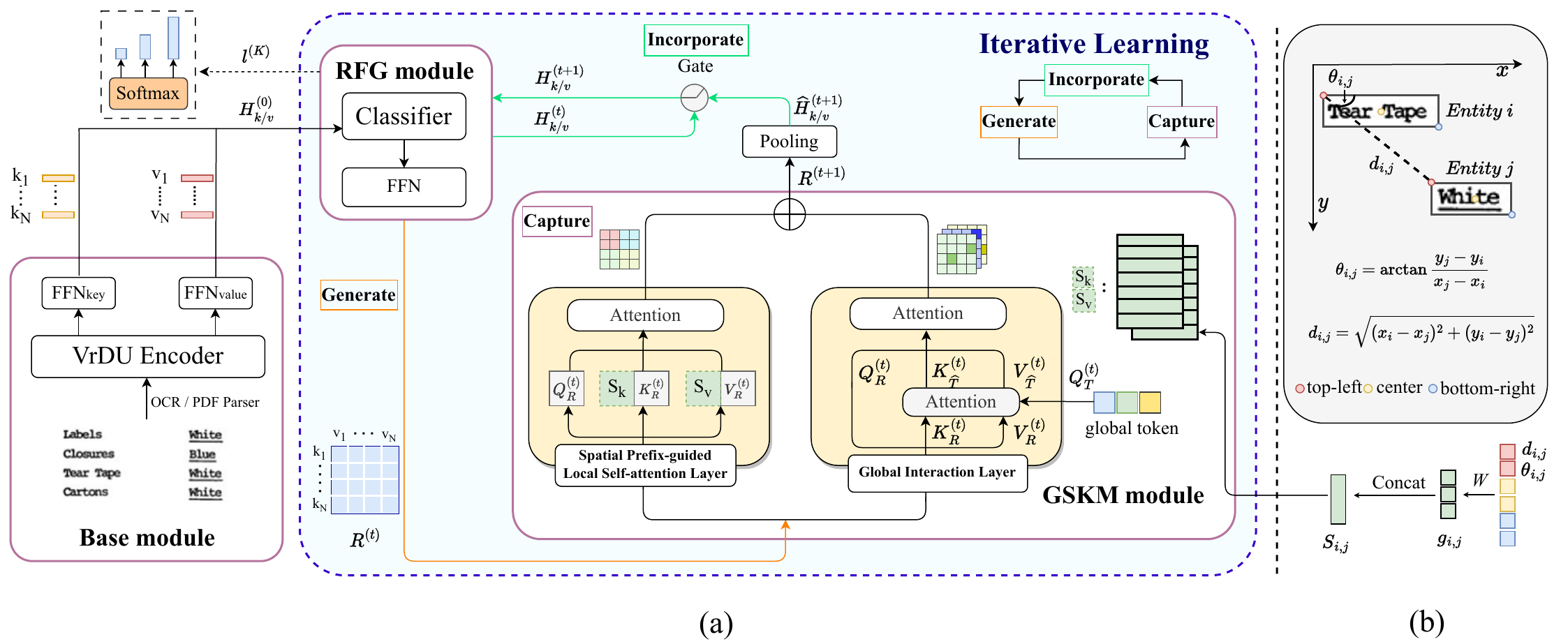}
    \caption{An illustration of the proposed {\model} framework. \textbf{(a)} The overview of {\model}. The framework consists of a Base module to generate initial \emph{key}/\emph{value} features based on entity representations from a pre-trained VrDU model, a relation feature generation (RFG) module to generate a relation feature for each entity pair, and a global structure knowledge mining (GSKM) module to mine global structure information in a local-global way. Besides, the RFG and GSKM modules are performed $K$ times in an iterative way to generate final predictions. \textbf{(b)} Spatial prefix construction. The spatial geometric features of entity pairs are computed as the spatial prefix to guide attention.}
    \label{overview}
\end{figure*}
In this section, we first formalize the visually-rich document relation extraction task and then briefly introduce how the task was approached in the past. 

\subsection{Problem Formulation}
The input to the VRE task is a scanned image of a document. Each visually rich document contains a set of semantic entities, and each entity is composed of a group of words and coordinates of the bounding box. We use a lower-case letter $e$ to represent semantic entity, where $e = \{[w^1,w^2,...,w^k], [x^1,y^1,x^2,y^2]\}$. The sequence $[w^1,w^2,...,w^k]$ means the word group, $x^1/x^2$ and $y^1/y^2$ are left/right x-coordinates and top/down y-coordinates respectively. The corresponding boldface lower-case letter $\textbf{e}$ indicates its embedding.
Let $\mathcal{E}$ and $\mathcal{R}$ represent the set of entities and relations respectively, where $\mathcal{E} = \{{e_i\}_{i=1}^N}$, $\mathcal{R} = \{(e_i, e_j) \}\subseteq \mathcal{E} \times \mathcal{E}$, $(e_i, e_j)$ mean the key-value entity pair and the directed link from $e_i$ to $e_j$. So each visually rich document can be denoted as $\mathcal{D} = \{ \mathcal{E}, \mathcal{R}\}$. The goal of VRE task is to determine whether a relation (link) exists between any two semantic entities. Notably, the semantic entity may exist relations with multiple entities or does not have relations with any other entities.

\subsection{Fine-tuning for Visual Relation Extraction}
Inspired by the success of pre-training in visually-rich document understanding, most existing methods~\cite{funsd,xu2021layoutxlm,lilt} fine-tune pre-trained VrDU model for VRE task. These methods take entity representations from a pre-trained VrDU model as input and then train a binary classifier to predict relations for all possible semantic entity pairs. Specifically, these methods project entity representations to \emph{key}/\emph{value} features respectively by two FFN layers. The features of \emph{key} and \emph{value} are concatenated and fed into a binary classifier to compute loss.


\section{Methodology}
\label{sec:method}


In this section, we describe 
the proposed framework named {\model} in detail. As shown in Figure~\ref{overview}, our method consists of three main modules, a Base module, a Relation Feature Generation (RFG) module, and a Global Structure Knowledge Mining (GSKM) module. In Section~\ref{method:base}, we present how the Base module can be combined with a pre-trained VrDU model to generate initial \emph{key}/\emph{value} features. In Section~\ref{method:rfg}, we introduce the RFG module to generate relation features. Then, in Section~\ref{method:gskm}, we elaborate on how the GSKM module mines global structure knowledge in a local-global way. We further provide a theoretical analysis to better understand the efficient and effective features of the GSKM module. Finally, we show how to apply our iterative learning strategy to mine accurate global structure knowledge in Section~\ref{method:iterative}.



\subsection{Base Module}
\label{method:base}
Given a visually rich document, the Base module first generates preliminary \emph{key}/\emph{value} features, which is same as most fine-tuning methods~\cite{xu2021layoutxlm,xylayoutlm,lilt}. Specifically, we use a pre-trained VrDU model to obtain semantic entity representations. Then entity representation \emph{\textbf{e}} is fed into two separated \emph{Feed-Forward Networks} (\emph{FFN}) to generate the initial \emph{key} feature and \emph{value} feature (denoted as $H_k^{(0)}$ and $H_v^{(0)}$ respectively), as written with Eq.~\eqref{eq:Encoder}:
\begin{equation}
\begin{aligned}
&H_k^{(0)}=W_{key}\textbf{e}+b_{key},\\
&H_v^{(0)}=W_{value}\textbf{e}+b_{value},  
\label{eq:Encoder}
\end{aligned}
\end{equation}
where $W_{key/value} \in \mathbb{R}^{2d_h \times d_h}$ are trainable  weights and $b_{key/value} \in \mathbb{R}^{d_h}$ are trainable biases. 

\subsection{Relation Feature Gneration (RFG) Module}
\label{method:rfg}
Taking \emph{key}/\emph{value} features as input, the RFG module generates relation features for all entity pairs. We denote \emph{key}/\emph{value} features at the $t$-$th$ iteration as $H_k^{(t)}$ and $H_v^{(t)}$ respectively, where $H_{k/v}^{(t)} \in \mathbb{R}^{N \times d_h}$. At the t-th iteration, the RFG module concatenates $H_k^{(t)}$ and $H_v^{(t)}$ as input and uses a bi-affine classifier to compute the relation \emph{logits} of each entity pairs with Eq.\eqref{eq:logits}:
\begin{equation}
l^{(t)} = H_{k}^{(t)}W_1H_{v}^{(t)} + H_{k}^{(t)}W_2,
\label{eq:logits}\end{equation}
where $l^{(t)}\in \mathbb{R}^{N \times N \times 2}$ denotes relation \emph{logits} at the $t$-$th$ iteration. Then we employ a \emph{FFN} block to generate a unified relation feature map (denoted as $R^{(t)} \in \mathbb{R}^{N \times N \times d_h}$), which contains relation features of all entity pairs. the relation feature map is calculated as follows:
\begin{equation}
\begin{aligned}
&R^{(t)}=W_{r}{l^{(t)}}+b_{r}, 
\label{eq:rel-feature}
\end{aligned}
\end{equation}
where $W_{r} \in \mathbb{R}^{2 \times d_h}$ is trainable  weight and $b_{r} \in \mathbb{R}^{d_h}$ is trainable biases.

\subsection{Global Structure Knowledge Mining (GSKM) module}
\label{method:gskm}
The GSKM module mines the global structure knowledge on the relation feature map in a local-global way. As illustrated in Figure~\ref{overview}, this module consists of a spatial prefix-guided local self-attention (SPLS) layer and a global interaction layer. The SPLS layer first partitions the relation feature map into multiple non-overlapping windows and then performs spatial prefix-guided local self-attention independently in each window.
The global interaction layer brings long-range dependencies to the local self-attention with negligible computation cost.

\subsubsection{Spatial Prefix-guided Local Self-attention (SPLS) Layer}
To address the challenge of mining global structure knowledge in a large computational space, we consider the spatial layout information of entity pairs in VrDs to guide self-attention computation. Most of existing VrDU methods independently encode the spatial information of each entity in the pre-training phase (i.e., the coordinate information of the bounding boxes in 2D space). However, it is necessary to insert spatial layout information of entity pairs explicitly in the fine-tuning phase (i.e., the spatial information of linking lines between bounding boxes in 2D space). Therefore, inspired by the success of prompt learning~\cite{prefix,li2023gradient} we propose a spatial prefix-guided local self-attention (SPLS) layer to mine global structure knowledge in a parameter-efficient manner.

\noindent\textbf{Spatial Prefix Construction.}
We calculate spatial geometric features (denoted as $S \in \mathbb{R}^{N\times N\times d_h}$) of linking lines between each entity pair as the spatial prefix. As shown in Figure~\ref{overview}(b), for an entity pair $(e_i, e_j)$, we calculate the direction and Euclidean distance of the line linking from the vertice $(x_i, y_i)$ of the bounding box of $e_i$ to the same vertice $(x_j, y_j)$ of $e_j$ as follows:
\begin{equation}
\begin{aligned}
&g_{i,j} = [W_{\theta}\theta(i,j);W_{d}d(i,j)],\\
&d(i,j) = \sqrt{(x_i-x_j)^2+(y_i-y_j)^2}, \\
&\theta(i,j) = \arctan{\frac{y_j-y_i}{x_j-x_i}},
\label{eq:geo-feature}
\end{aligned}
\end{equation}
where $W_{\theta/d} \in \mathbb{R}^{1\times \frac{d_h}{6}}$ are trainable weights. 
Therefore the spatial geometric features $S_{i,j}$ of entity pair $(e_i,e_j)$ is calculated as follows:
\begin{equation}
\begin{aligned}
&S_{i,j}=[g_{i,j}^{tl}; g_{i,j}^{ct}; g_{i,j}^{br}],
\label{eq:rel-geo}
\end{aligned}
\end{equation}
where $g^{tl}$, $g^{ct}$, $g^{br}$ indicate top-left, center, and bottom-right points respectively. Then we treat $S$ as the spatial prefix and compute attention on the hybrid keys and values.

\noindent\textbf{Spatial Prefix-guided Attention.} We first partition $R^{(t)}$ into non-overlapping windows and then perform spatial prefix-guided local attention within each local window. The variant formula of self-attention with the spatial prefix as follows \footnote{Without loss of generalization, we ignore the scaling factor $\sqrt{d}$ of the softmax operation for the convenience of explanation.}:
\begin{equation}
\label{eq:spatial-prefix}
\small
\begin{split}
& \mR^{(t)}_{local}  = \text{softmax}\big( \mR^{(t)}\mW_q [SW_k^s; \mR^{(t)}\mW_k] ^\top\big) \begin{bmatrix}  SW_v^s \\ \mR^{(t)}\mW_v \end{bmatrix} \\
&=\text{softmax}(\mQ_R^{(t)}[S_k; \mK_R^{(t)}]^\top) \begin{bmatrix} S_v \\ \mV_R^{(t)}\end{bmatrix}\\
& = (1 - \lambda(\textbf{R}^{(t)})) \underbrace{ \text{Attn}(\mQ_R^{(t)}, \mK_R^{(t)}, \mV_R^{(t)})}_{\text{standard attention}} \\
&+ \lambda(\textbf{R}^{(t)}) \underbrace{ \text{Attn}((\mQ_R^{(t)}, S_k,S_v)}_{\text{spatial-prefix guided attention}},
\end{split}
\end{equation}

\begin{equation}
\small
\lambda(\bm{R}^{(t)}) = \frac{\sum_i\exp (\mQ_R^{(t)} S_k^{\top})_i}{\sum_i\exp (\mQ_R^{(t)} S_k^{\top})_i + \sum_j \exp(\mQ_R^{(t)} \mK_R^{{(t)}^{\top}})_j},
\end{equation} 
where $R^{(t)}_{local}$ refers to the local attention output at the t-th iteration, $\lambda(\bm{R}^{(t)})$ denotes the scalar for the sum of normalized attention weights on the key and value vectors from spatial prefix.

\subsubsection{Global Interaction Layer (GIL)}
After the SPLS layer effectively aggregates local correlations with window priors, we introduce a global interaction layer to bring long-range dependencies to the local self-attention. As illustrated in Figure~\ref{overview}, we use learnable global tokens $T \in \mathbb{R}^{M\times d_h}$ to compute the global interaction at the t-th iteration as follows:
\begin{equation}
\begin{aligned}
&\widehat{T}^{(t)} = \text{Attn}(Q_T^{(t)}, \mK_R^{(t)}, \mV_R^{(t)}),\\
&\mR_{global}^{(t)} = \text{Attn}(\mQ_R^{(t)},K_{\widehat{T}}^{(t)},V_{\widehat{T}}^{(t)}),
\label{eq:global}
\end{aligned}
\end{equation} 
where $R^{(t)}_{global}$ refers to the global interaction at the t-th iteration. $T$ will be updated in the same way as $H_{k/v}$ throughout the iterative learning process.
Subsequently, we compute $R^{(t+1)}$ 
and employ the mean pooling operation to obtain the context-aware key and value features as:
\begin{equation}
\begin{aligned}
&\mR^{(t+1)} = \mR_{local}^{(t)} + \mR_{global}^{(t)},\\
&\widehat{H}_{k/v}^{(t+1)} = \text{mean-pooling}(\mR^{(t+1)}),
\label{pool}
\end{aligned}
\end{equation}
where $\widehat{H}_{k/v}^{(t+1)}$ contain global structure information.

\noindent\textbf{Analysis of GSKM.} Here we give some analysis to help better understand GSKM, especially effective and efficient features. \\
\noindent\textbf{Effectiveness.} GSKM can effectively learn global structure knowledge guided by spatial layout information in VrDs. As shown in Eq.~\ref{eq:spatial-prefix}, 
the first term $\text{Attn}(\mQ_R^{(t)}, \mK_R^{(t)}, \mV_R^{(t)})$ is the standard attention in the content side, whereas the second term represents the 2D spatial layout guidelines. In this sense, our method implements 2D spatial layout to guide the attention computation in a way similar to linear interpolation. Specifically, the GSKM module down-weights the original content attention probabilities by a scalar factor (i.e., $1-\lambda$) and redistributes the remaining attention probability $\lambda$ to attend to spatial-prefix guided attention, which likes the linear interpolation.

\noindent\textbf{Efficiency.} GSKM can reduce the computation complexity $N^4$ to $N^2\times S^2$, where $S$ denotes window size. In the SPLS layer, with a relation feature map $R^{(t)}\in \mathbb{R}^{N \times N \times d_h}$ as input, we partition $R^{(t)}$ into non-overlapping windows with shape $(\frac{N}{S}\times\frac{N}{S}, S\times S,d_h )$ to reduce the computation complexity $N^4$ of self-attention to $(\frac{N}{S}\times\frac{N}{S})\times (S\times S)^2= N^2\times S^2$, where $S$ denotes window size. Meanwhile, the computation complexity of the global interaction layer ($N^2 \times M$) is negligible, as the number of global tokens $M$ is much smaller than the window size $S^2$ in our method.

\subsection{Iterative Learning}
\label{method:iterative}
To alleviate the noisy mining process, we further propose an iterative learning strategy to enable global structure information and entity embeddings mutually reinforce each other. Specifically, we incorporate global structure knowledge into entity representations through a gating mechanism:
\begin{equation}
\label{gate}
\begin{aligned}
& g = \text{sigmoid}(W_g[H_{k/v}^{(t)};\widehat{H}_{k/v}^{(t+1)}]+b_g)\\
&H_{k/v}^{(t+1)} = H_{k/v}^{(t)} + g \cdot \widehat{H}_{k/v}^{(t+1)}
\end{aligned}
\end{equation} 
Finally, these new key and value features are fed back to the classifier for the next iteration. After repeating this iterative process $K$ times, we get updated key and value features $H_{k/v}^{(K)}$ to compute final logits $l^{(K)}$. Finally, 
we calculate Binary Cross Entropy (BCE) loss based on $l^{(K)}$ as follows
\footnote{To make a fair comparison with baselines, we use the same binary classification training strategy.}:
\begin{equation}
 \label{bce_loss}
\mathcal{L}=\sum_{i=1}^{N} \sum_{j=1}^{N} \ell(l_{i,j}^{(K)}, y_{i,j})
\end{equation}
where $y_{i,j} \in [0,1]$ is binary ground truth of the entity pair $(e_i,e_j)$, $\ell(.,.)$ is the cross-entropy loss. 




\section{Experiments}
\label{sec:experiment}
\begin{table*}[htbp!]
\centering
\scalebox{0.72}{
\begin{tabular}{c|l|c|ccccccc|c}
\toprule
\multirow{2}{*}{Structure Information}&
 \multicolumn{1}{c|}{\multirow{2}{*}{Model}} &FUNSD & \multicolumn{7}{c|}{XFUND}    & \multirow{2}{*}{Avg.} \\ 
 \cmidrule(l){3-10}                       
 && EN    & ZH & JA & ES & FR & IT & DE & PT &   
 \\ \midrule \midrule
 \multirow{2}{*}{Text-only} 
 &XLM-RoBERTa            &  0.2659 &  0.5105  & 0.5800  & 0.5295 &  0.4965  & 0.5305 &  0.5041 &  0.3982  & 0.4769  \\
&InfoXLM              &  0.2920 &  0.5214  & 0.6000  & 0.5516 &  0.4913  & 0.5281 &  0.5262  & 0.4170  & 0.4910    \\
\midrule
\multirow{3}{*}{Local} 
&LayoutXLM                 &  0.5483  & 0.7073  & 0.6963 &  0.6896 &  0.6353  & 0.6415 &  0.6551 &  0.5718  & 0.6432 \\
&XYLayoutLM&-&0.7445&0.7059&0.7259&0.6521&0.6572&0.6703&0.5898&- \\
 &LiLT                   &  0.6276                      &   0.7297                 &     0.7037               &        0.7195             &    0.6965               &          0.7043           &         0.6558            &      0.5874              &  0.6781 \\

\midrule
\multirow{2}{*}{Global} 
&{\model}$_{LayoutXLM}$&0.5926&0.8631&\textbf{0.8258}&0.8375&0.7729&0.8035&0.7780&0.7026&0.7720\tiny{\color{red}{(+12.88\%)}} \\
&{\model}$_{LiLT}$&\textbf{0.7697}&\textbf{0.8752}&0.8096&\textbf{0.8595}&\textbf{0.8646}&\textbf{0.8415}&\textbf{0.8023}&\textbf{0.7384}&\textbf{0.8201} \tiny{\color{red}{(+14.20\%)}} \\
\bottomrule
\end{tabular}}
\caption{Language-specific fine-tuning F1 accuracy on FUNSD and XFUND (fine-tuning on X, testing on X)."Text-only" denotes pre-trained textual models without structure information, "Local" denotes pre-trained VrDU models with local features, and "Global" denotes using global structure information.}
\label{tab:language-specific}
\end{table*}
\begin{table*}[htbp!]
\centering
\scalebox{0.72}{
\begin{tabular}{c|l|c|ccccccc|c}
\toprule
\multirow{2}{*}{Structure Information}&
 \multicolumn{1}{c|}{\multirow{2}{*}{Model}} &FUNSD & \multicolumn{7}{c|}{XFUND}    & \multirow{2}{*}{Avg.} \\ 
 \cmidrule(l){3-10}                       
 && EN    & ZH & JA & ES & FR & IT & DE & PT &   
 \\ \midrule \midrule
 \multirow{2}{*}{Text-only} 
&XLM-RoBERTa                   &  0.2659 &0.1601& 0.2611 &0.2440 &0.2240 &0.2374 &0.2288 &0.1996& 0.2276 \\
&InfoXLM                    &  0.2920& 0.2405 &0.2851 &0.2481 &0.2454& 0.2193 &0.2027 &0.2049 &0.2423  \\
\midrule
\multirow{2}{*}{Local} 
&LayoutXLM                 &  0.5483& 0.4494& 0.4408 &0.4708 &0.4416 &0.4090 &0.3820& 0.3685 &0.4388  \\
&LiLT                   &    0.6276                  &     0.4764               &    0.5081                 &     0.4968            & 0.5209               &   0.4697             &   0.4169                &  0.4272          &  0.4930  \\
\midrule
\multirow{2}{*}{Global} 
&\model$_{LayoutXLM}$&0.5926&0.5696&0.5556&0.5124&0.5295&0.4168&0.4325&0.4363&0.5056 \tiny{\color{red}{(+6.68\%)}}\\
&\model$_{LiLT}$&\textbf{0.7697}&\textbf{0.6930}&\textbf{0.6805}&\textbf{0.7072}&\textbf{0.7145}&\textbf{0.6355}&\textbf{0.5997}&\textbf{0.5830}&\textbf{0.6729} \tiny{\color{red}{(+17.99\%)}} \\

\bottomrule 
\end{tabular}}
\caption{Cross-lingual zero-shot transfer F1 accuracy on FUNSD and XFUND (fine-tuning on FUNSD, testing on X).}
\label{tab:zero}
\end{table*}

In this section, we perform detailed experiments to demonstrate the effectiveness of our proposed method {\model} among different settings. Besides the standard setting of typical language-specific fine-tuning (section~\ref{sec:standard}), we further consider more challenging settings to demonstrate the generalizability of {\model} such as cross-lingual zero-shot transfer learning (section~\ref{sec:cross}) and few-shot learning (section~\ref{sec:low}). Before discussing the results, we provide the details of the experimental setup below.
\subsection{Experimental Setup}
\subsubsection{Datasets}
\noindent \textbf{FUNSD}~\cite{funsd} is a scanned document dataset for form understanding. It has 149 training samples and 50 test samples with various layouts. \\
\noindent \textbf{XFUND} \cite{xu2021layoutxlm} is a multilingual form understanding benchmark. It includes 7 languages with 1,393 fully annotated forms. Each language includes 199 forms. where the training set includes 149 forms, and the test set includes 50 forms.
\subsubsection{Baselines}
We use the following baselines: (1) text-only pre-trained models without structure information: XLM-RoBERT \cite{xlm-r}, InfoXLM \cite{chi2021infoxlm}; (2) layout-aware pre-trained VrDU models with local structure information: LayoutXLM \cite{xu2021layoutxlm}, XYLayoutLM \cite{xylayoutlm}, LiLT \cite{lilt}. All of the experimental results of these baselines are from their original papers directly, except for the results on the few-shot learning task~\footnote{We re-implement the results using the official code.}.
\subsubsection{Experiment Implementation}
We use the bi-affine classifier and binary classification training loss over all datasets and settings following~\cite{lilt} for a fair comparison. The entity representation is the first token vector in each entity. For the few-shot learning task, we randomly sample training samples over each shot five times with different random seeds, and report the average performance under five sampling times for a fair comparison. More details of the training
hyper-parameters can be found in Appendix~\ref{app:hyper}.

\subsection{Language-specific Fine-tuning}
\label{sec:standard}
We compare the performance of {\model} applied to
the language-specific fine-tuning task. The experimental results are shown in Table~\ref{tab:language-specific}. First, all VRE methods with structure information outperform the text-only models XLM-RoBERT and InfoXLM, which indicates structure information plays an important role in the VrDU. Second, while pre-trained VrDU models have achieved significant improvement over text-only models, our method still outperforms them by a large margin. This phenomenon denotes that incorporating global structure information is generally helpful for VRE. Compared to the SOTA method LiLT~\cite{lilt}, {\model} achieves significant improvements on all language datasets and has an increase of 14.20\% F1 accuracy on the average performance. Third, we further observe that our {\model} is model-agnostic, which can consistently improves diverse pre-trained models’ relation extraction performance on all datasets. For example, {\model} has an improvement of 12.88\% F1 accuracy on average performance compared to LayoutXLM~\cite{xu2021layoutxlm}.



\subsection{Cross-lingual Transfer Learning}
\label{sec:cross}

We evaluate {\model} on the cross-lingual zero-shot transfer learning task. In this setting, the model is only fine-tuned on the FUNSD dataset (in English) and evaluated on each specific language dataset. We present the evaluation results in Table~\ref{tab:zero}. It can be observed that {\model} significantly outperforms its competitors and consistently improve diverse backbone encoders’ relation extraction performance on all datasets. This verifies that {\model} can capture the common global structure information invariance among different languages and transfer it to other languages for VRE. We observe that the performance improvement of {\model}(LayoutXLM) is not as significant as {\model}(LiLT). This may be attributed to that the architecture of LiLT decoupling the text and layout information makes it easier to learn language-independent structural information, which is consistent with the observation of previous works~\cite{lilt}. We further evaluate GOSE on the Multilingual learning task, the results are shown in Appendix~\ref{app:multi}.





\subsection{Few-shot Learning}
\label{sec:low}
\begin{figure}[t]
    \centering
    \includegraphics[height=3.3cm]{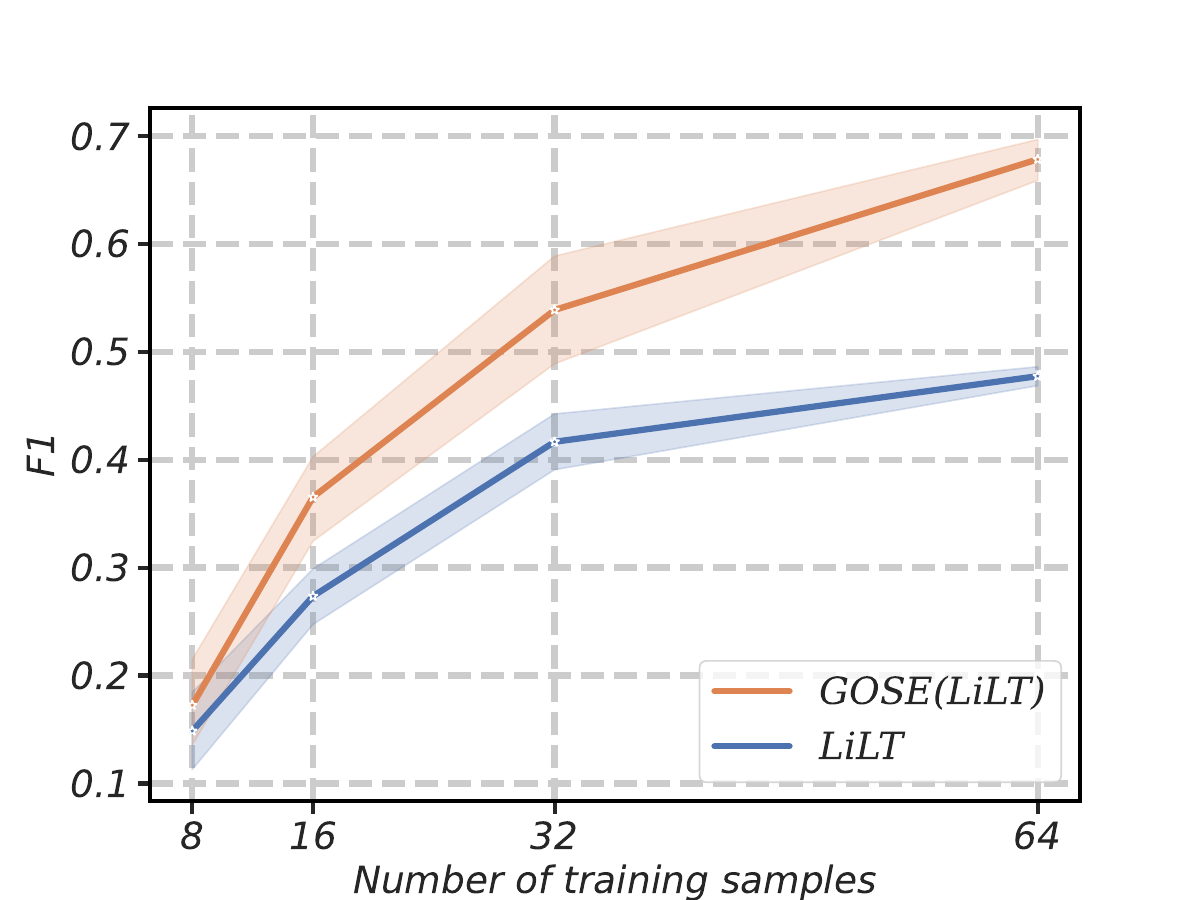}
    \caption{Average performances of few-shot setting on FUNSD dataset. We report the results of random sampling five times over each shot.}
    \label{fig:few}
\end{figure} 

Previous experiments illustrate that our method achieves improvements using full training samples. We further explore whether {\model} could mine global structure information in the low-resource setting. Thus, we compare with the previous SOTA model LiLT on few-shot settings. The experimental results in Figure~\ref{fig:few} indicate that the average performance of {\model} still outperforms the SOTA model LiLT. Notably, our {\model} achieves a comparable average performance (67.82\%) on FUNSD dataset using only 64 training samples than LiLT does (62.76\%) using full training samples, which further proves that our proposed method can more efficiently leverage training samples. This success may be attributed to the incorporation of global structure knowledge can improve the generalization of the model.





\begin{table}[t]
\centering
\resizebox{0.48\textwidth}{!}{\begin{tabular}{clccccc}
\toprule
&\multicolumn{1}{c}{\multirow{2}{*}{Method}} & \multicolumn{3}{c}{Components} &\multicolumn{2}{c}{F1 Accuracy$\uparrow$}\\ 
\cmidrule(l){3-5}  \cmidrule(l){6-7} 
&  &GIL&Spatial-Prefix&GSKM & EN&ZH \\ 
\midrule \midrule
  & {\model}  & \Checkmark & \Checkmark& \Checkmark  &  \textbf{0.7697}& \textbf{0.8752}\\ 
 \midrule
 1 & w/o GIL  & \XSolidBrush & \Checkmark& \Checkmark &  0.7384& 0.8643\\
 2 & w/o Spatial-Prefix   & \Checkmark & \XSolidBrush & \Checkmark &  0.7029& 0.8161\\ 
 3 & w/o GSKM   & \XSolidBrush& \XSolidBrush & \XSolidBrush  & 0.6902 &0.8037 \\ 
\bottomrule
\end{tabular}}
\caption{Ablation study of our model using LiLT as the backbone on the FUNSD and XUND (ZH). The symbol EN denotes FUNSD and ZH means chinese language.}
\label{tab:ablation}
\end{table}

\subsection{Ablation Study}
\noindent \textbf{Effectiveness of individual components.} We further investigate the effectiveness of different modules in our method. we compare our model with the following variants in Table~\ref{tab:ablation}.

    

(1) \emph{w/o GIL}. In this variant, we remove the global interaction layer from GSKM.  This change means that the GSKM module only performs local self-attention. The results shown in Table~\ref{tab:ablation} suggest that our GIL can encourage the GSKM module to better exploit dependencies of entity pairs.

(2) \emph{w/o Spatial-Prefix}.  In this variant, we remove the spatial prefix-guided interaction. This change causes a significant performance decay. This suggests the injection of additional spatial layout information can guide the attention mechanism attending to entity pairs with similar spatial structure and thus help the model learn global structural knowledge. \\
\begin{figure*}[htb]
    \begin{minipage}{1.0\linewidth}
        \centering
        \centerline{\includegraphics[height=4.1cm]{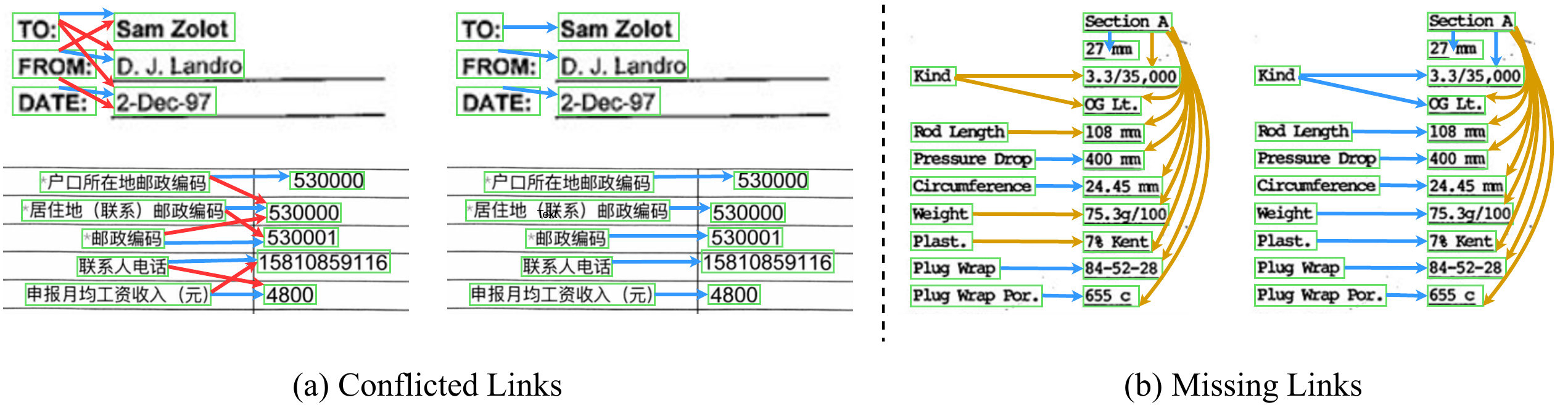}}
    \end{minipage}%
	\caption{The visualization of examples in FUNSD and XFUND(ZH). The left of each example is LiLT and the right is {\model}(LiLT). The arrows in blue, red and orange denote true positive, false positive and false negative (missed) relations respectively. Best viewed by zooming up.}
	\label{fig:visual} 
\end{figure*}
(3) \emph{w/o GSKM}. In this variant, we remove the GSKM module from GOSE. This change means the model only obtains context-aware relation features through a mean-pooling operation. The results shown in Table~\ref{tab:ablation} indicate that although the mean-pooling operation can achieve a performance improvement, the GSKM module can mine more useful global structure information.


\begin{figure}[htbp]
\small
\begin{minipage}[t]{0.24\textwidth}
\centering
\includegraphics[height=2.8cm]{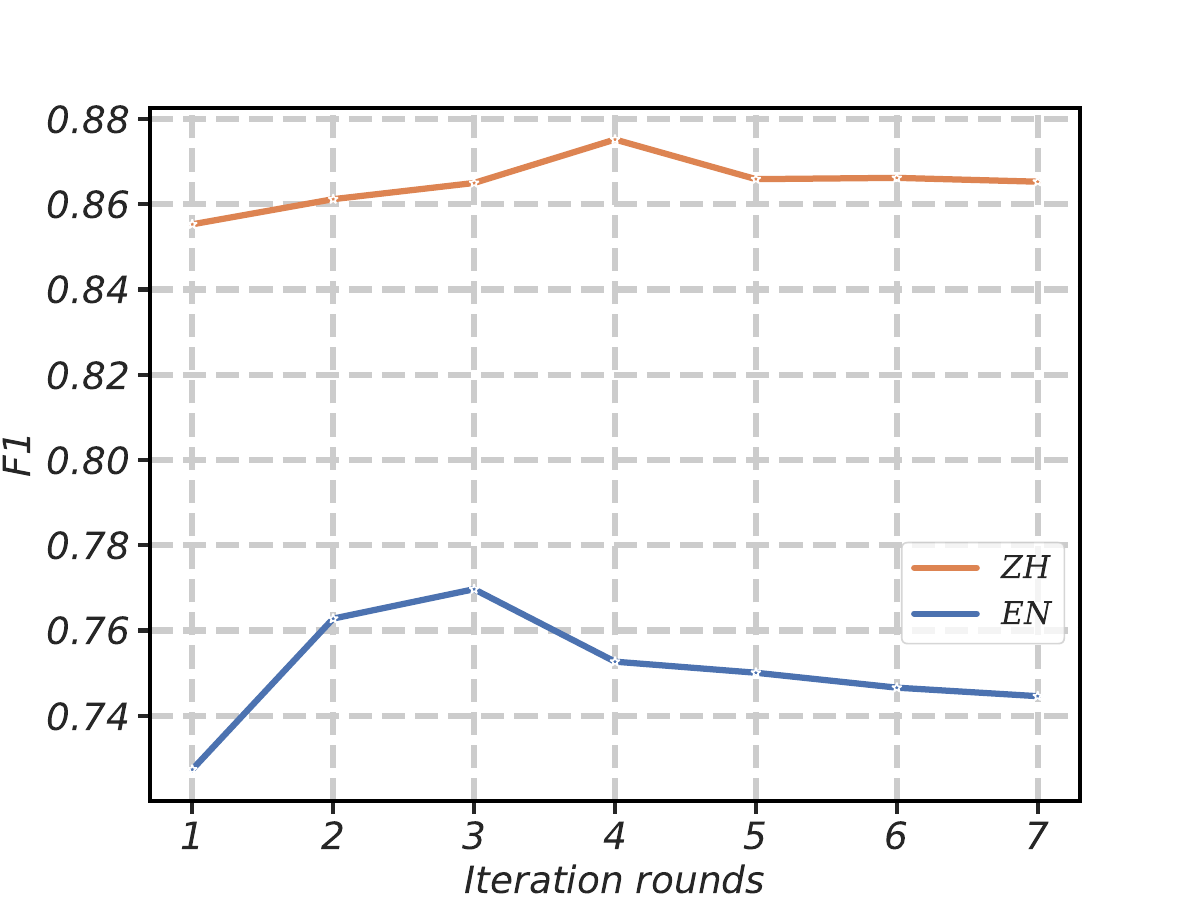}\\
\quad\quad (a)
\end{minipage}
\begin{minipage}[t]{0.2\textwidth}
\centering
\includegraphics[height=2.8cm]{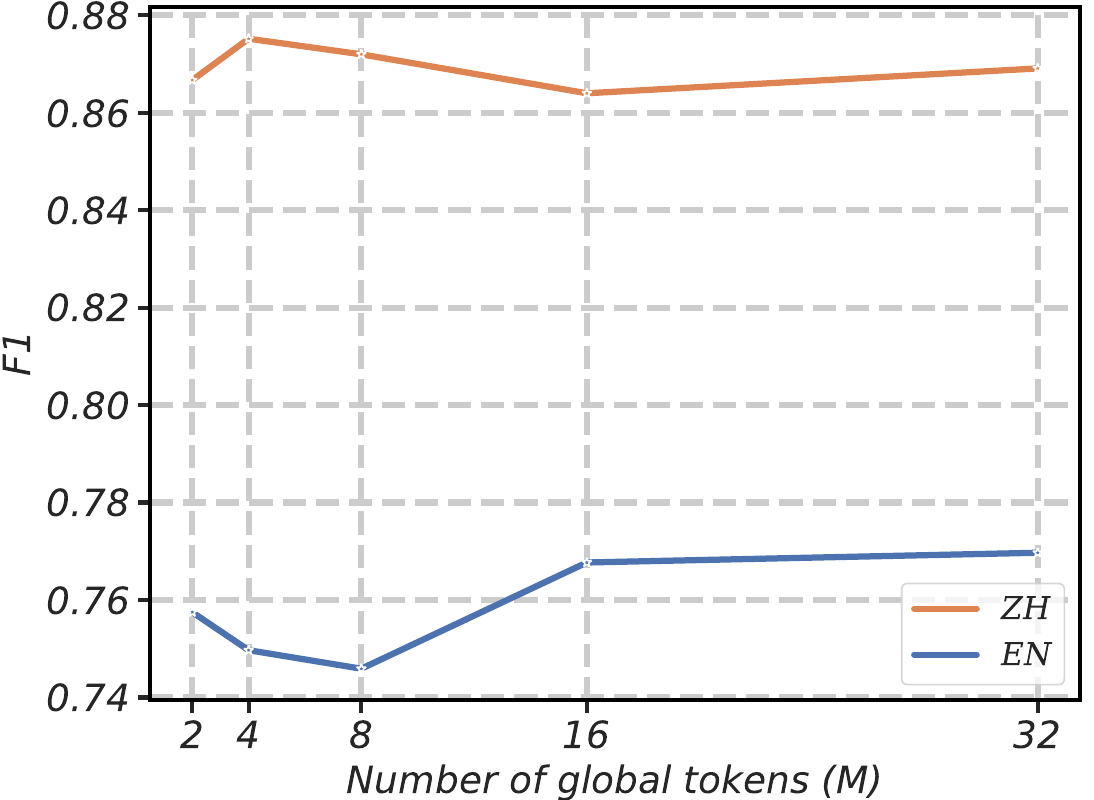}\\
\quad \quad\quad \small(b)
\end{minipage}
\caption{F1 performances of our model using LiLT as the
backbone. The symbol
EN denotes FUNSD dataset and ZH means XFUND(in chinese). \textbf{(a)} Ablation of iteration rounds. \textbf{(b)} Ablation of global tokens.}
\label{fig:rounds}
\end{figure}
\noindent \textbf{Ablation of Iteration Rounds.} The highlight of our {\model} is mining global structure knowledge and refining entity embeddings iteratively. We argue that these two parts can mutually reinforce each other: the integration of global structure knowledge can help refine entity embeddings. On the contrary, better entity embeddings can help mine more accurate global structure knowledge. Thus, we evaluate the influence of the iteration rounds. The results are shown in ~\ref{fig:rounds}(a), which indicates GOSE usually achieves the best results within 
small number of iteration rounds. In addition, we investigate the performance of different information under multiple iterations in Appendix \ref{app:iteration}.

\noindent \textbf{Ablation of Global Tokens.}
We further investigate the effect of the number of global tokens in the GIL on our model. The results are shown in Figure~\ref{fig:rounds} (b), which denotes {\model} can achieve optimal results within a small number of global tokens, while keeping efficiency.

\subsection{Further Analysis}
\noindent \textbf{Case Study.}
To better illustrate the effectiveness of global structure knowledge, we conduct the specific case analysis on the VRE task as shown in Figure~\ref{fig:visual}. Through the visualization of examples, we can notice: 
(1) as shown in Figure~\ref{fig:visual}(a), {\model} can greatly mitigate the prediction of conflicted links which reveals that our method can capture global structure knowledge to detect conflicted interactions between entity pairs.
(2) as shown in Figure~\ref{fig:visual}(b), {\model} can learn long-range relations by analogy with the linking pattern of entity pairs, while keeping a good recall. Notably, it is also difficult for {\model} to predict long-range relations where is not sufficient global structure knowledge. For example, {\model} does not predict well relations of the entity "\textit{Section A}", due to there are few top-to-bottom and one-to-many linking patterns. 

\begin{figure}[ht]
    \centering
    \includegraphics[width=0.48\textwidth]{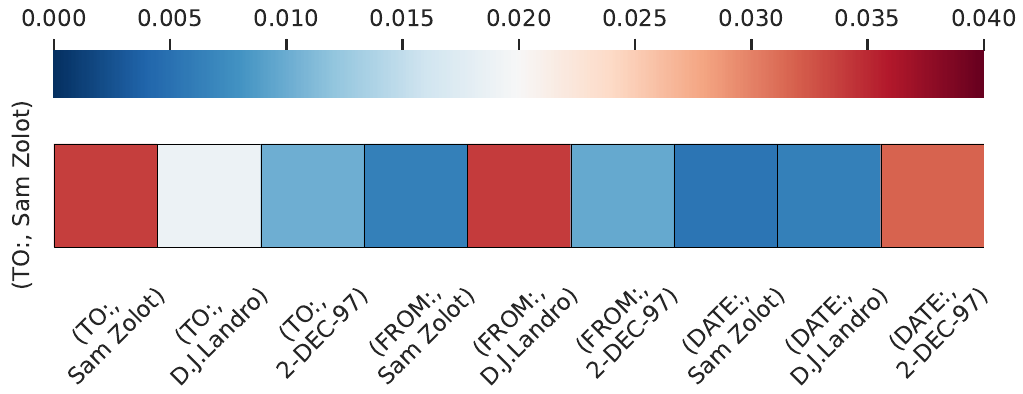}
    \caption{The attention weight visualization between the entity pair (TO:, Sam Zolot) and other spatial prefix of entity pairs.}
    \label{atten_weight}
\end{figure}

\noindent \textbf{Visualization of Attention over Spatial Information.}
To illustrate the effect
of our proposed spatial prefix-guided local self-attention. We calculate the attention scores over the spatial information for the document in Figure 5(a), i.e., the spatial-prefix guided attention weights using Equation (6). As shown in Figure~\ref{atten_weight}, the entity pair (TO:, Sam Zolot) pays more attention towards the entity pair (FROM:, D.J.Landro) and (DATE:, 2-DEC-97). This phenomenon indicates that the injection of additional spatial layout information of entity pairs can guide the attention mechanism attending to entity pairs with similar spatial structure, thereby enhancing the capacity of the model to discern precise dependencies between entity pairs.
\section{Related Works}
\subsection{Visual Relation Extraction}
Visual relation extraction (VRE) aims at identifying relations between semantic entities from visually-rich documents.
Early methods based on graph neural networks~\cite{sera, tang2021matchvie,li2021adaptive,li2022compositional,li2023variational} learned node features by treating semantic entities as nodes in a graph. Recently, most studies~\cite{li2020unsupervised,lilt,xylayoutlm,huang2022layoutlmv3},
used the self-supervised pre-training and fine-tuning techniques to boost the performance on document understanding. Although they have achieved significant improvement on VIE tasks, especially on the semantic entity recognition (SER) task~\cite{ernie-layout}. VRE remains largely underexplored and is also a challenging task.
In this paper, we focus on mining global structure knowledge to guide relation extraction. To the best of our knowledge, our work is the first attempt to exploit dependencies of entity pairs for this task.

\subsection{Efficient Transformers}
Efficient transformers~\cite{dong2019unified,li2019walking,efficient,longformer,video,zhang2021poolingformer,li2022dilated} are a class of methods designed to address the quadratic time and memory complexity problems of vanilla self-attention. More similar to us are methods that leverage down-sampling to reduce the resolution of the sequence, such as window-based vision transformers~\cite{swin,focal,lightvit,uformer}.
Different from existing methods, we propose the spatial prefix-guided attention mechanism, which leverages spatial layout properties of VrDs to guide {\model} to mine global structure knowledge.

\section{Discussion}
Recently, The Multimodal Large Language Model (MLLM)~\cite{li2022fine,yin2023survey} has emerged as a pioneering approach. MLLMs leverage powerful Large Language Models (LLMs) as a brain to perform multimodal tasks. The surprising emergent capabilities of MLLM, such as following zero-shot demonstrative instructions~\cite{li2023finetuning} and OCR-free math reasoning~\cite{zhu2023minigpt4,dai2023instructblip}, are rare in traditional methods. Several studies~\cite{xu2023lvlmehub,liu2023hidden} have conducted comprehensive evaluations of publicly available large multimodal models. These investigations reveal that MLLMs still struggle with the VIE task. In this paper, we introduce and empirically validate that global structural knowledge is useful for visually-rich document information extraction. Our insights have the potential to shape the advancement of large model technology in the domain of visually-rich documents.

\section{Conclusion}
In this paper, we present a general
global structure knowledge-guided relation extraction method for visually-rich documents, which jointly and iteratively learns entity representations and mines global dependencies of entity pairs. To the best of our knowledge, {\model} is the first work leveraging global structure knowledge to guide visual relation extraction. Concretely, we first use a base module that combines with a pre-trained model to obtain initial relation predictions. Then, we further design a relation feature generation module that generates relation features and a global structure knowledge mining module. These two modules perform the "generate-capture-incorporate” process multiple times to mine and integrate accurate global structure knowledge. Extensive experimental results on three different settings (e.g., standard fine-tuning, cross-lingual learning, low-resource setting) over eight datasets demonstrate the effectiveness and superiority of our {\model}.
\section*{Acknowledgments}
We would like to express gratitude to the anonymous reviewers for their kind comments. This work has been supported in part by the Zhejiang NSF (LR21F020004), Key Research and Development Projects in Zhejiang Province (No. 2023C01030, 2023C01032), NSFC (No. 62272411), National Key Research and Development Program of China (2018AAA0101900), Ant Group and Alibaba-Zhejiang University Joint Research Institute of Frontier Technologies.
Our work was also supported by Scientific Research Fund of Zhejiang Provincial Education Department.
\section*{Limitations}
The proposed work still contains several limitations to address in future work, as follows:

\noindent \textbf{Method.} One limitation of our method is that it cannot 
capture global structure information from informative visual
clues. The visual features in visually-rich documents such as font size and color may provide diverse structure information. For example, section titles in resumes and job ads are often in fonts different from the content. We leave this for future work.

\noindent \textbf{Task.} We only evaluate the visual relation extraction task covering diverse settings. Due to the limited budget and computation resources, we cannot afford evaluation on more tasks related to visually-rich documents. We will plan to evaluate the proposed approach on more visual information extraction tasks such as semantic entity recognition.

\bibliography{citation}

\begin{thebibliography}{40}
\expandafter\ifx\csname natexlab\endcsname\relax\def\natexlab#1{#1}\fi

\bibitem[{Beltagy et~al.(2020)Beltagy, Peters, and Cohan}]{longformer}
Iz~Beltagy, Matthew~E. Peters, and Arman Cohan. 2020.
\newblock Longformer: The long-document transformer.
\newblock \emph{arXiv:2004.05150}.

\bibitem[{Chi et~al.(2021)Chi, Dong, Wei, Yang, Singhal, Wang, Song, Mao,
  Huang, and Zhou}]{chi2021infoxlm}
Zewen Chi, Li~Dong, Furu Wei, Nan Yang, Saksham Singhal, Wenhui Wang, Xia Song,
  Xian-Ling Mao, He-Yan Huang, and Ming Zhou. 2021.
\newblock {InfoXLM}: An information-theoretic framework for cross-lingual
  language model pre-training.
\newblock In \emph{NAACL-HLT}, pages 3576--3588.

\bibitem[{Conneau et~al.(2020)Conneau, Khandelwal, Goyal, Chaudhary, Wenzek,
  Guzm{\'a}n, Grave, Ott, Zettlemoyer, and Stoyanov}]{xlm-r}
Alexis Conneau, Kartikay Khandelwal, Naman Goyal, Vishrav Chaudhary, Guillaume
  Wenzek, Francisco Guzm{\'a}n, {\'E}douard Grave, Myle Ott, Luke Zettlemoyer,
  and Veselin Stoyanov. 2020.
\newblock Unsupervised cross-lingual representation learning at scale.
\newblock In \emph{ACL}, pages 8440--8451.

\bibitem[{Cui et~al.(2021)Cui, Xu, Lv, and Wei}]{doc-ai}
Lei Cui, Yiheng Xu, Tengchao Lv, and Furu Wei. 2021.
\newblock Document {AI:} benchmarks, models and applications.
\newblock \emph{arXiv preprint arXiv:2111.08609}.

\bibitem[{Dai et~al.(2023)Dai, Li, Li, Tiong, Zhao, Wang, Li, Fung, and
  Hoi}]{dai2023instructblip}
Wenliang Dai, Junnan Li, Dongxu Li, Anthony Meng~Huat Tiong, Junqi Zhao,
  Weisheng Wang, Boyang Li, Pascale Fung, and Steven Hoi. 2023.
\newblock \href {http://arxiv.org/abs/2305.06500} {Instructblip: Towards
  general-purpose vision-language models with instruction tuning}.

\bibitem[{Dong et~al.(2019)Dong, Yang, Wang, Wei, Liu, Wang, Gao, Zhou, and
  Hon}]{dong2019unified}
Li~Dong, Nan Yang, Wenhui Wang, Furu Wei, Xiaodong Liu, Yu~Wang, Jianfeng Gao,
  Ming Zhou, and Hsiao-Wuen Hon. 2019.
\newblock Unified language model pre-training for natural language
  understanding and generation.
\newblock \emph{Proceedings of NeurIPS}.

\bibitem[{Gu et~al.(2022)Gu, Meng, Wang, Lan, Wang, Gu, and Zhang}]{xylayoutlm}
Zhangxuan Gu, Changhua Meng, Ke~Wang, Jun Lan, Weiqiang Wang, Ming Gu, and
  Liqing Zhang. 2022.
\newblock Xylayoutlm: Towards layout-aware multimodal networks for
  visually-rich document understanding.
\newblock In \emph{Proceedings of the IEEE/CVF Conference on Computer Vision
  and Pattern Recognition (CVPR)}, pages 4583--4592.

\bibitem[{He et~al.(2022)He, Zhou, Ma, Berg{-}Kirkpatrick, and Neubig}]{prefix}
Junxian He, Chunting Zhou, Xuezhe Ma, Taylor Berg{-}Kirkpatrick, and Graham
  Neubig. 2022.
\newblock Towards a unified view of parameter-efficient transfer learning.

\bibitem[{Huang et~al.(2022{\natexlab{a}})Huang, Huang, You, Wang, Qian, and
  Xu}]{lightvit}
Tao Huang, Lang Huang, Shan You, Fei Wang, Chen Qian, and Chang Xu.
  2022{\natexlab{a}}.
\newblock Lightvit: Towards light-weight convolution-free vision transformers.
\newblock \emph{arXiv preprint arXiv:2207.05557}.

\bibitem[{Huang et~al.(2022{\natexlab{b}})Huang, Lv, Cui, Lu, and
  Wei}]{huang2022layoutlmv3}
Yupan Huang, Tengchao Lv, Lei Cui, Yutong Lu, and Furu Wei. 2022{\natexlab{b}}.
\newblock Layoutlmv3: Pre-training for document ai with unified text and image
  masking.
\newblock In \emph{Proceedings of the 30th ACM International Conference on
  Multimedia}.

\bibitem[{Hwang et~al.(2021)Hwang, Yim, Park, Yang, and Seo}]{spatial}
Wonseok Hwang, Jinyeong Yim, Seunghyun Park, Sohee Yang, and Minjoon Seo. 2021.
\newblock Spatial dependency parsing for semi-structured document information
  extraction.
\newblock In \emph{Findings of ACL}.

\bibitem[{Jaume et~al.(2019)}]{funsd}
Guillaume Jaume et~al. 2019.
\newblock {FUNSD}: A dataset for form understanding in noisy scanned documents.
\newblock In \emph{ICDAR}, volume~2, pages 1--6.

\bibitem[{Li et~al.(2021{\natexlab{a}})Li, Bi, Yan, Wang, Huang, Huang, and
  Si}]{li2021structurallm}
Chenliang Li, Bin Bi, Ming Yan, Wei Wang, Songfang Huang, Fei Huang, and Luo
  Si. 2021{\natexlab{a}}.
\newblock {StructuralLM}: Structural pre-training for form understanding.
\newblock In \emph{ACL}.

\bibitem[{Li et~al.(2023{\natexlab{a}})Li, Gao, Wei, Tang, Zhang, Li, Ji, Tian,
  Chua, and Zhuang}]{li2023gradient}
Juncheng Li, Minghe Gao, Longhui Wei, Siliang Tang, Wenqiao Zhang, Mengze Li,
  Wei Ji, Qi~Tian, Tat-Seng Chua, and Yueting Zhuang. 2023{\natexlab{a}}.
\newblock Gradient-regulated meta-prompt learning for generalizable
  vision-language models.
\newblock In \emph{Proceedings of the IEEE/CVF International Conference on
  Computer Vision}.

\bibitem[{Li et~al.(2022{\natexlab{a}})Li, He, Wei, Qian, Zhu, Xie, Zhuang,
  Tian, and Tang}]{li2022fine}
Juncheng Li, Xin He, Longhui Wei, Long Qian, Linchao Zhu, Lingxi Xie, Yueting
  Zhuang, Qi~Tian, and Siliang Tang. 2022{\natexlab{a}}.
\newblock Fine-grained semantically aligned vision-language pre-training.
\newblock \emph{Advances in neural information processing systems},
  35:7290--7303.

\bibitem[{Li et~al.(2023{\natexlab{b}})Li, Pan, Ge, Gao, Zhang, Ji, Zhang,
  Chua, Tang, and Zhuang}]{li2023finetuning}
Juncheng Li, Kaihang Pan, Zhiqi Ge, Minghe Gao, Hanwang Zhang, Wei Ji, Wenqiao
  Zhang, Tat-Seng Chua, Siliang Tang, and Yueting Zhuang. 2023{\natexlab{b}}.
\newblock Fine-tuning multimodal llms to follow zero-shot demonstrative
  instructions.
\newblock \emph{arXiv preprint arXiv:2308.04152}.

\bibitem[{Li et~al.(2019)Li, Tang, Wu, and Zhuang}]{li2019walking}
Juncheng Li, Siliang Tang, Fei Wu, and Yueting Zhuang. 2019.
\newblock Walking with mind: Mental imagery enhanced embodied qa.
\newblock In \emph{Proceedings of the 27th ACM International Conference on
  Multimedia}, pages 1211--1219.

\bibitem[{Li et~al.(2021{\natexlab{b}})Li, Tang, Zhu, Shi, Huang, Wu, Yang, and
  Zhuang}]{li2021adaptive}
Juncheng Li, Siliang Tang, Linchao Zhu, Haochen Shi, Xuanwen Huang, Fei Wu,
  Yi~Yang, and Yueting Zhuang. 2021{\natexlab{b}}.
\newblock Adaptive hierarchical graph reasoning with semantic coherence for
  video-and-language inference.
\newblock In \emph{Proceedings of the IEEE/CVF International Conference on
  Computer Vision}, pages 1867--1877.

\bibitem[{Li et~al.(2023{\natexlab{c}})Li, Tang, Zhu, Zhang, Yang, Chua, and
  Wu}]{li2023variational}
Juncheng Li, Siliang Tang, Linchao Zhu, Wenqiao Zhang, Yi~Yang, Tat-Seng Chua,
  and Fei Wu. 2023{\natexlab{c}}.
\newblock Variational cross-graph reasoning and adaptive structured semantics
  learning for compositional temporal grounding.
\newblock \emph{IEEE Transactions on Pattern Analysis and Machine
  Intelligence}.

\bibitem[{Li et~al.(2020)Li, Wang, Tang, Shi, Wu, Zhuang, and
  Wang}]{li2020unsupervised}
Juncheng Li, Xin Wang, Siliang Tang, Haizhou Shi, Fei Wu, Yueting Zhuang, and
  William~Yang Wang. 2020.
\newblock Unsupervised reinforcement learning of transferable meta-skills for
  embodied navigation.
\newblock In \emph{Proceedings of the IEEE/CVF Conference on Computer Vision
  and Pattern Recognition}, pages 12123--12132.

\bibitem[{Li et~al.(2022{\natexlab{b}})Li, Xie, Qian, Zhu, Tang, Wu, Yang,
  Zhuang, and Wang}]{li2022compositional}
Juncheng Li, Junlin Xie, Long Qian, Linchao Zhu, Siliang Tang, Fei Wu, Yi~Yang,
  Yueting Zhuang, and Xin~Eric Wang. 2022{\natexlab{b}}.
\newblock Compositional temporal grounding with structured variational
  cross-graph correspondence learning.
\newblock In \emph{Proceedings of the IEEE/CVF Conference on Computer Vision
  and Pattern Recognition}, pages 3032--3041.

\bibitem[{Li et~al.(2022{\natexlab{c}})Li, Xie, Zhu, Qian, Tang, Zhang, Shi,
  Zhang, Wei, Tian, and Zhuang}]{li2022dilated}
Juncheng Li, Junlin Xie, Linchao Zhu, Long Qian, Siliang Tang, Wenqiao Zhang,
  Haochen Shi, Shengyu Zhang, Longhui Wei, Qi~Tian, and Yueting Zhuang.
  2022{\natexlab{c}}.
\newblock Dilated context integrated network with cross-modal consensus for
  temporal emotion localization in videos.
\newblock In \emph{Proceedings of the 30th ACM International Conference on
  Multimedia}.

\bibitem[{Liu et~al.(2023)Liu, Li, Li, Yu, Liu, Yang, Huang, Peng, Liu, Chen,
  Li, Yin, lin Liu, Jin, and Bai}]{liu2023hidden}
Yuliang Liu, Zhang Li, Hongliang Li, Wenwen Yu, Yang Liu, Biao Yang, Mingxin
  Huang, Dezhi Peng, Mingyu Liu, Mingrui Chen, Chunyuan Li, Xucheng Yin, Cheng
  lin Liu, Lianwen Jin, and Xiang Bai. 2023.
\newblock \href {http://arxiv.org/abs/2305.07895} {On the hidden mystery of ocr
  in large multimodal models}.

\bibitem[{Liu et~al.(2021)Liu, Lin, Cao, Hu, Wei, Zhang, Lin, and Guo}]{swin}
Ze~Liu, Yutong Lin, Yue Cao, Han Hu, Yixuan Wei, Zheng Zhang, Stephen Lin, and
  Baining Guo. 2021.
\newblock Swin transformer: Hierarchical vision transformer using shifted
  windows.
\newblock In \emph{Proceedings of the IEEE/CVF International Conference on
  Computer Vision (ICCV)}, pages 10012--10022.

\bibitem[{Loshchilov and Hutter(2018)}]{adamw}
Ilya Loshchilov and Frank Hutter. 2018.
\newblock Decoupled weight decay regularization.
\newblock In \emph{ICLR}.

\bibitem[{Paszke et~al.(2019)Paszke, Gross, Massa, Lerer, Bradbury, Chanan,
  Killeen, Lin, Gimelshein, Antiga, Desmaison, K{\"{o}}pf, Yang, DeVito,
  Raison, Tejani, Chilamkurthy, Steiner, Fang, Bai, and Chintala}]{pytorch}
Adam Paszke, Sam Gross, Francisco Massa, Adam Lerer, James Bradbury, Gregory
  Chanan, Trevor Killeen, Zeming Lin, Natalia Gimelshein, Luca Antiga, Alban
  Desmaison, Andreas K{\"{o}}pf, Edward~Z. Yang, Zachary DeVito, Martin Raison,
  Alykhan Tejani, Sasank Chilamkurthy, Benoit Steiner, Lu~Fang, Junjie Bai, and
  Soumith Chintala. 2019.
\newblock Pytorch: An imperative style, high-performance deep learning library.
\newblock In \emph{Proceedings of the Annual Conference on Neural Information
  Processing Systems}, pages 8024--8035.

\bibitem[{Peng et~al.(2022)Peng, Pan, Wang, Luo, Zhang, Huang, Cao, Yin, Chen,
  Zhang, Feng, Sun, Tian, Wu, and Wang}]{ernie-layout}
Qiming Peng, Yinxu Pan, Wenjin Wang, Bin Luo, Zhenyu Zhang, Zhengjie Huang,
  Yuhui Cao, Weichong Yin, Yongfeng Chen, Yin Zhang, Shikun Feng, Yu~Sun, Hao
  Tian, Hua Wu, and Haifeng Wang. 2022.
\newblock Ernie-layout: Layout knowledge enhanced pre-training for
  visually-rich document understanding.
\newblock In \emph{Findings of the Association for Computational Linguistics:
  EMNLP 2022}, pages 3744--3756.

\bibitem[{Ryoo et~al.(2021)Ryoo, Piergiovanni, Arnab, Dehghani, and
  Angelova}]{video}
Michael Ryoo, AJ~Piergiovanni, Anurag Arnab, Mostafa Dehghani, and Anelia
  Angelova. 2021.
\newblock Tokenlearner: Adaptive space-time tokenization for videos.
\newblock In \emph{Advances in Neural Information Processing Systems}, pages
  12786--12797.

\bibitem[{Tang et~al.(2021)Tang, Xie, Jin, Wang, Chen, Xu, Wang, Wu, and
  Li}]{tang2021matchvie}
Guozhi Tang, Lele Xie, Lianwen Jin, Jiapeng Wang, Jingdong Chen, Zhen Xu,
  Qianying Wang, Yaqiang Wu, and Hui Li. 2021.
\newblock {MatchVIE}: Exploiting match relevancy between entities for visual
  information extraction.
\newblock In \emph{IJCAI}, pages 1039--1045.

\bibitem[{Tay et~al.(2020)Tay, Dehghani, Bahri, and Metzler}]{efficient}
Yi~Tay, Mostafa Dehghani, Dara Bahri, and Donald Metzler. 2020.
\newblock \href {http://arxiv.org/abs/2009.06732} {Efficient transformers: {A}
  survey}.
\newblock \emph{CoRR}, abs/2009.06732.

\bibitem[{Wang et~al.(2022{\natexlab{a}})Wang, Jin, and Ding}]{lilt}
Jiapeng Wang, Lianwen Jin, and Kai Ding. 2022{\natexlab{a}}.
\newblock {L}i{LT}: A simple yet effective language-independent layout
  transformer for structured document understanding.
\newblock In \emph{Proceedings of the 60th Annual Meeting of the Association
  for Computational Linguistics (ACL)}, pages 7747--7757.

\bibitem[{Wang et~al.(2022{\natexlab{b}})Wang, Cun, Bao, Zhou, Liu, and
  Li}]{uformer}
Zhendong Wang, Xiaodong Cun, Jianmin Bao, Wengang Zhou, Jianzhuang Liu, and
  Houqiang Li. 2022{\natexlab{b}}.
\newblock Uformer: A general u-shaped transformer for image restoration.
\newblock In \emph{Proceedings of the IEEE/CVF Conference on Computer Vision
  and Pattern Recognition (CVPR)}, pages 17683--17693.

\bibitem[{Xu et~al.(2023)Xu, Shao, Zhang, Gao, Liu, Lei, Meng, Huang, Qiao, and
  Luo}]{xu2023lvlmehub}
Peng Xu, Wenqi Shao, Kaipeng Zhang, Peng Gao, Shuo Liu, Meng Lei, Fanqing Meng,
  Siyuan Huang, Yu~Qiao, and Ping Luo. 2023.
\newblock \href {http://arxiv.org/abs/2306.09265} {Lvlm-ehub: A comprehensive
  evaluation benchmark for large vision-language models}.

\bibitem[{Xu et~al.(2020)Xu, Li, Cui, Huang, Wei, and Zhou}]{Layoutlm}
Yiheng Xu, Minghao Li, Lei Cui, Shaohan Huang, Furu Wei, and Ming Zhou. 2020.
\newblock {LayoutLM}: Pre-training of text and layout for document image
  understanding.
\newblock In \emph{ACM-SIGKDD}, pages 1192--1200.

\bibitem[{Xu et~al.(2021)Xu, Lv, Cui, Wang, Lu, Florencio, Zhang, and
  Wei}]{xu2021layoutxlm}
Yiheng Xu, Tengchao Lv, Lei Cui, Guoxin Wang, Yijuan Lu, Dinei Florencio, Cha
  Zhang, and Furu Wei. 2021.
\newblock {LayoutXLM}: Multimodal pre-training for multilingual visually-rich
  document understanding.
\newblock \emph{arXiv preprint arXiv:2104.08836}.

\bibitem[{Yang et~al.(2021)Yang, Li, Zhang, Dai, Xiao, Yuan, and Gao}]{focal}
Jianwei Yang, Chunyuan Li, Pengchuan Zhang, Xiyang Dai, Bin Xiao, Lu~Yuan, and
  Jianfeng Gao. 2021.
\newblock Focal attention for long-range interactions in vision transformers.
\newblock In \emph{Advances in Neural Information Processing Systems}.

\bibitem[{Yin et~al.(2023)Yin, Fu, Zhao, Li, Sun, Xu, and Chen}]{yin2023survey}
Shukang Yin, Chaoyou Fu, Sirui Zhao, Ke~Li, Xing Sun, Tong Xu, and Enhong Chen.
  2023.
\newblock \href {http://arxiv.org/abs/2306.13549} {A survey on multimodal large
  language models}.

\bibitem[{Zhang et~al.(2021{\natexlab{a}})Zhang, Gong, Shen, Li, Lv, Duan, and
  Chen}]{zhang2021poolingformer}
Hang Zhang, Yeyun Gong, Yelong Shen, Weisheng Li, Jiancheng Lv, Nan Duan, and
  Weizhu Chen. 2021{\natexlab{a}}.
\newblock Poolingformer: Long document modeling with pooling attention.
\newblock \emph{Proceedings of ICML}.

\bibitem[{Zhang et~al.(2021{\natexlab{b}})Zhang, Bo, Wang, Cao, Li, and
  Bao}]{sera}
Yue Zhang, Zhang Bo, Rui Wang, Junjie Cao, Chen Li, and Zuyi Bao.
  2021{\natexlab{b}}.
\newblock Entity relation extraction as dependency parsing in visually rich
  documents.
\newblock In \emph{Proceedings of the 2021 Conference on Empirical Methods in
  Natural Language Processing}, pages 2759--2768.

\bibitem[{Zhu et~al.(2023)Zhu, Chen, Shen, Li, and Elhoseiny}]{zhu2023minigpt4}
Deyao Zhu, Jun Chen, Xiaoqian Shen, Xiang Li, and Mohamed Elhoseiny. 2023.
\newblock \href {http://arxiv.org/abs/2304.10592} {Minigpt-4: Enhancing
  vision-language understanding with advanced large language models}.

\end{thebibliography}
\clearpage
\bibliographystyle{acl_natbib}
\appendix
\section{Hyperparameters}
\label{app:hyper}
All of our experiments are performed on one NVIDIA 3090 GPU with the PyTorch~\cite{pytorch} framework. We use a mini-batch AdamW~\cite{adamw} optimizer with a weight decay of 0.1. The model is trained with a batch size of 2. The window size is fixed to 64. In the Language-Specific Fine-tuning experiments for all languages, the learning rate, steps are set to $6.25\times10^{-6}$,$2\times10^{4}$ accordingly on LiLT encoder, while $2.5\times10^{-5}$,$4\times10^{4}$ on the Layoutxlm encoder. In the multilingual learning experiments, we use the full language-specific data for training, with total steps $1.6\times10^{5}$. In the cross-lingual zero-shot transfer learning experiments, we directly evaluate the model, which was trained in the previous language-specific experiments, on the XFUND dataset.

\section{Multilingual Learning}
\label{app:multi}
We evaluate {\model} on the multilingual learning setting. In this setting, the model is fine-tuned with all 8 languages simultaneously and evaluated on each specific language. From the experimental results shown in Table~\ref{tab:multitask}, we can find that although this setting further improves the baseline model performance compared to the language-specific fine-tuning, our method {\model} once again outperforms its counterparts by a large margin. We hold that the superior performance of our method {\model} can be attributed to the fact that the previous method still does not learn sufficient global structure information in multilingual learning. This finding also demonstrates that mining global structure information is beneficial for the VRE task.

\begin{table}[ht]
\centering
\resizebox{0.48\textwidth}{!}{\begin{tabular}{c|l|c|c}
\toprule
 & Model&FUNSD&XFUND (Avg.) \\ 
\midrule \midrule
 \multirow{2}{*}{Text-only}
&XLM-RoBERTa                 &   0.3638 & 0.6727 \\
&InfoXLM                    &    0.3699 & 0.6495\\
\midrule
\multirow{2}{*}{Local}
&LayoutXLM & 0.6671 & 0.7988 \\
&LiLT      & 0.7407 & 0.8228\\
\midrule
\multirow{2}{*}{Global}
&\model$_{LayoutXLM}$&0.7755& 0.8656\\
&{\model}$_{LiLT}$&\textbf{0.9003}&\textbf{0.8994} \\
\bottomrule 
\end{tabular}}
\caption{ Multilingual fine-tuning F1 accuracy on FUNSD and XFUND (fine-tuning on 8 languages all, testing on X).}
\label{tab:multitask}
\end{table}

\section{Effect of Different Information}
\label{app:iteration}
\begin{figure}[t]
    \centering
    \small
   \includegraphics[width=0.65\linewidth]{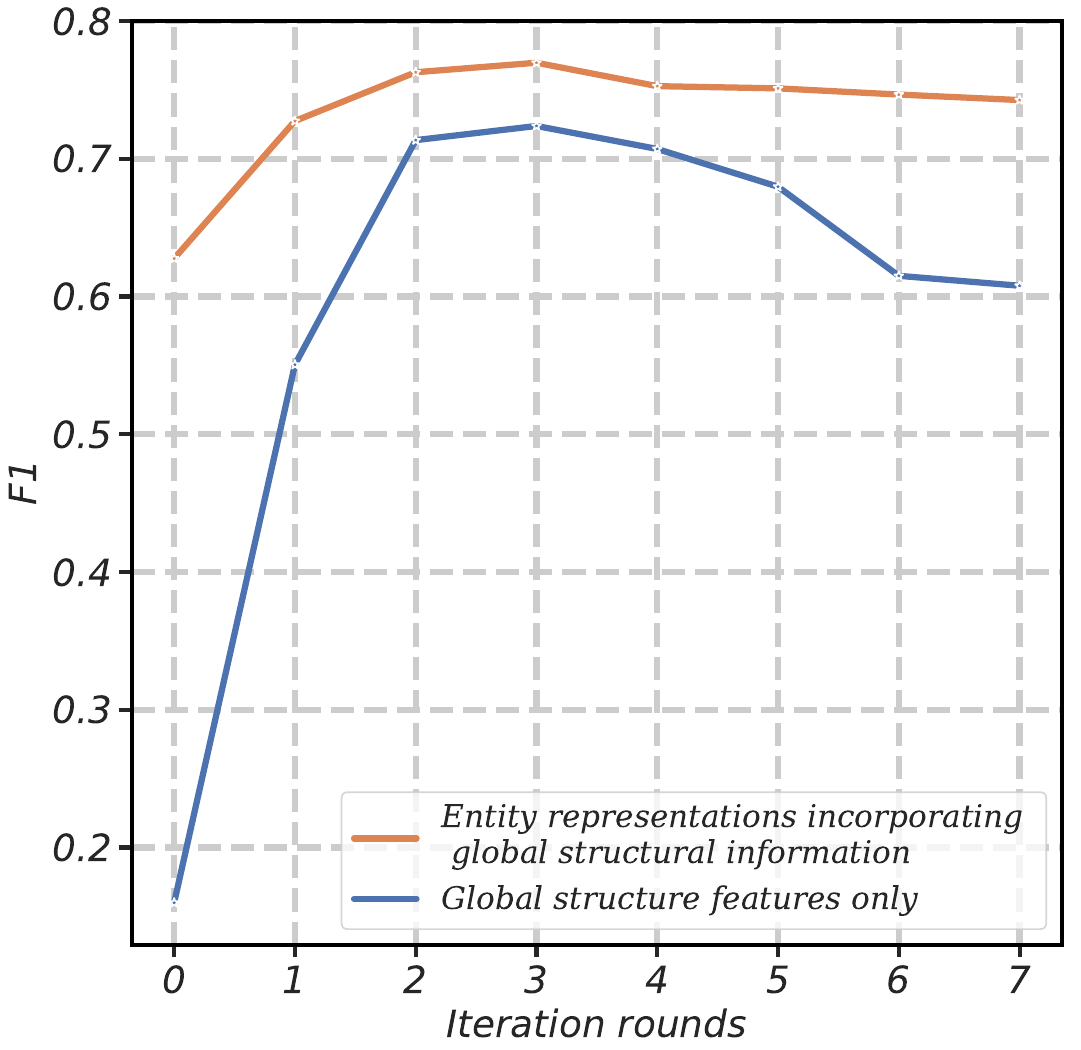}
    \caption{F1 performances of different information of our GOSE (LiLT) under multiple iterations on FUNSD dataset.}
    \label{fig:information}
\end{figure}

We show the performance of our GOSE (LiLT) under multiple iterations in Figure~\ref{fig:information}. We can observe that (1) when the iterative learning process begins, both global structural features and entity representations progressively enhance and mutually reinforce each other. (2)  
In scenarios where the number of iteration rounds stands at zero, i.e., without iterative learning. The performance of global structure information is poor. This may be because the initial entity representations obtained from the pre-trained model are not strong, thus the mined structural information without iterative optimization is noisy.
(3) As the number of iteration rounds increases to a certain point, the performance of the model decreases. This phenomenon can be attributed to too many iteration rounds that can cause mined global structural information to become over-smoothing thus affecting the final performance of our model.

\end{document}